\documentclass[11pt,a4paper]{article}
\usepackage[hyperref]{AACL-IJCNLP2020}
\usepackage{times}
\usepackage{latexsym}
\usepackage{graphicx}
\usepackage{multirow}
\usepackage{hyperref}
\usepackage{fdsymbol}
\usepackage{xcolor}
\usepackage{booktabs}
\usepackage{placeins}

\usepackage{microtype}

\aclfinalcopy 
\def\aclpaperid{162} 

\title{Happy Are Those Who Grade without Seeing: A Multi-Task Learning Approach to Grade Essays Using Gaze Behaviour}

\author{Sandeep Mathias$^{\spadesuit}$, Rudra Murthy$^{\spadesuit,\diamondsuit}$, Diptesh Kanojia$^{\spadesuit,\clubsuit}$, Abhijit Mishra$^{\diamondsuit}$, Pushpak Bhattacharyya$^{\spadesuit}$\\
  $^{\spadesuit}$ Department of Computer Science, Indian Institute of Technology, Bombay \\
  $^{\diamondsuit}$ IBM Research, India \\
  $^{\clubsuit}$ IITB-Monash Research Academy \\
  \texttt{\{sam,rudra,diptesh,pb\}@cse.iitb.ac.in, abhijitmishra.530@gmail.com} \\}

\date{}

\begin{document}
\maketitle
\begin{abstract}
The gaze behaviour of a reader is helpful in solving several NLP tasks such as automatic essay grading. However, collecting gaze behaviour from readers is costly in terms of time and money. In this paper, we propose a way to improve automatic essay grading using gaze behaviour, which is learnt at run time using a multi-task learning framework. To demonstrate the efficacy of this multi-task learning based approach to automatic essay grading, we collect gaze behaviour for 48 essays across 4 essay sets, and learn gaze behaviour for the rest of the essays, numbering over 7000 essays. Using the learnt gaze behaviour, we can achieve a statistically significant improvement in performance over the state-of-the-art system for the essay sets where we have gaze data. We also achieve a statistically significant improvement for 4 other essay sets, numbering about 6000 essays, where we have no gaze behaviour data available. Our approach establishes that learning gaze behaviour improves automatic essay grading.

\end{abstract}

\section{Introduction}
\label{Introduction Section}

Collecting a reader's psychological input can be very beneficial to a number of Natural Language Processing (NLP) tasks, like complexity  \cite{mishra-etal-2017-scanpath,gonzalez-garduno-sogaard-2017-gaze-readability}, sentence simplification \cite{klerke-etal-2016-improving}, text understanding \cite{mishra-etal-2016-sarcasm-understandability}, text quality \cite{mathias-etal-2018-eyes}, parsing \cite{hale-etal-2018-finding}, \textit{etc.} This psychological information can be extracted using devices like eye-trackers, and electroencephalogram (EEG) machines. However, one of the challenges in using reader's information involves collecting the psycholinguistic data itself.


In this paper, we choose the task of automatic essay grading and show how we can predict the score that a human rater would give using both text and \textbf{\textit{learnt}} gaze behaviour. An essay is a piece of text, written in response to a topic, called a prompt. Automatic essay grading is assigning a score to the essay using a machine. An essay set is a set of essays written in response to the same prompt.

Multi-task learning \cite{Caruana1998} is a machine learning paradigm where we utilize auxiliary tasks to aid in solving a primary task. This is done by exploiting similarities between the primary task and the auxiliary tasks. \textbf{Scoring the essay} is the \textit{primary task} and \textbf{learning gaze behaviour} is the \textit{auxiliary task}.

Using gaze behaviour for a very small number of essays \textbf{(less than 0.7\% of the essays in an essay set)}, we see an improvement in predicting the overall score of the essays. We also use our gaze behaviour dataset to run experiments on \textbf{unseen} essay sets - \textit{i.e.,} essay sets which have \textbf{no gaze behaviour data} - and observe improvements in the system's performance in automatically grading essays.

\paragraph{Contributions} The main contribution of our paper is describing how we use gaze behaviour information, in a multi-task learning framework, to automatically score essays outperforming the state-of-the-art systems. We will also release the gaze behaviour dataset\footnote{Gaze behaviour dataset: \url{http://www.cfilt.iitb.ac.in/cognitive-nlp/} \\ Essays: \url{https://www.kaggle.com/c/asap-aes}} and code\footnote{\url{https://github.com/lwsam/ASAP-Gaze}} - the first of its kind, for automatic essay grading - to facilitate further research in using gaze behaviour for automatic essay grading and other similar NLP tasks.

\subsection{Gaze Behaviour Terminology}
\label{Terminology Subsection}

An \textbf{\textit{Interest Area}} (IA) is an area of the screen that we are interested in. These areas are where some text is displayed, and not the white background on the left/right, as well as above/below the text. \textbf{Each word} is a separate and unique IA.

A \textbf{\textit{Fixation}} is an event when the reader's eye is focused on a part of the screen. For our experiments, we are concerned only with fixations that occur within the interest areas. Fixations that occur in the background are ignored.

A \textbf{\textit{Saccade}} is the path of the eye movement, as it goes from one fixation to the next. There are two types of saccades - Progressions and Regressions. \textbf{\textit{Progressions}} are saccades where the reader moves from the current interest area to a \textit{later} one. \textbf{\textit{Regressions}} are saccades where the reader moves from the current interest area to an \textit{earlier} one.

The rest of the paper is organized as follows. Section \ref{Motivation Section} describes our motivation for using eye-tracking and learning gaze behaviour from readers, over unseen texts. Section \ref{Related Work Section} describes some of the related work in the area of automatic essay grading, eye tracking and multi-task learning. Section \ref{Features Section} describes the gaze behaviour attributes used in our experiments, and the intuition behind them. We describe our dataset creation and experiment setup in Section \ref{Dataset Section}. In Section \ref{Results Section}, we report our results and present a detailed analysis. We present our conclusions and discuss possible future work in Section \ref{Conclusion Section}.

\section{Motivation}
\label{Motivation Section}


\newcite{mishra2018cognitively}, for instance, describe a lot of research in solving multiple problems in NLP using gaze behaviour of readers. \textbf{However}, most of their work involves collecting the gaze behaviour data first, and then splitting the data into training and testing data, before performing their experiments. While their work did show significant improvements over baseline approaches, across multiple NLP tasks, collecting the gaze behaviour data would be quite expensive, both in terms of time and money.




Therefore, we ask ourselves: ``\textbf{\textit{Can we learn gaze behaviour, using a small amount of seed data, to help solve an NLP task?}}'' In order to use gaze behaviour on a large scale, we need to be able to \textit{learn} it, since we can not ask a user to read texts every time we wish to use gaze behaviour data. 
\newcite{mathias-etal-2018-eyes} describe using gaze behaviour to predict how a reader would rate a piece of text (which is similar to our chosen application). 
Since they showed that gaze behaviour can help in predicting text quality, we use multi-task learning to simultaneously learn gaze behaviour information (auxiliary task) as well as score the essay (the primary task).
However, they \textbf{collect all their gaze behaviour data \textit{a priori}}, while \textbf{\textit{we try to learn the gaze behaviour of a reader}} and use what we learn from our system, for grading the essays. Hence, while they showed that gaze behaviour \textit{could} help in predicting how a reader would score a text, their approach requires a reader to \textbf{read the text}, while our approach does not do so, \textbf{\textit{during testing / deployment}}. 

\section{Related Work}
\label{Related Work Section}

\subsection{Automatic Essay Grading (AEG)}
The very first AEG system was proposed by \newcite{page1966imminence}. Since then, there have been a lot of other AEG systems (see \newcite{shermis2013handbook} for more details). In 2012, the Hewlett Foundation released a dataset called the Automatic Student Assessment Prize (ASAP) AEG dataset. The dataset contains about 13,000 essays across eight different essay sets. We discuss more about that dataset later.

With the availability of a large dataset, there has been a lot of research, especially using neural networks, in automatically grading essays - like using Long Short Term Memory (LSTM) Networks \cite{taghipour-ng-2016-neural,tay-2018-skipflow}, Convolutional Neural Networks (CNNs) \cite{dong-zhang-2016-automatic}, or both \cite{dong-etal-2017-attention}. \newcite{zhang-litman-2018-co} improve on the results of \newcite{dong-etal-2017-attention} using co-attention between the source article and the essay for one of the types of essay sets.

\subsection{Eye-Tracking}

Capturing the gaze behaviour of readers has been found to be quite useful in improving the performance of NLP tasks \cite{mishra2018cognitively}. The main idea behind using gaze behaviour is the eye-mind hypothesis \cite{just1980theory}, which states that whatever text the eye reads, that is what the mind processes. This hypothesis has led to a large body of work in psycholinguistic research that shows a relationship between text processing and gaze behaviour. \newcite{mishra2018cognitively} also describe some of the ways that eye-tracking can be used for multiple NLP tasks like translation complexity, sentiment analysis, etc.

Research has been done on using gaze behaviour at run time to solve downstream NLP tasks like sentence simplification \cite{klerke-etal-2016-improving}, readability \cite{gonzalez2018learning,singh-etal-2016-quantifying}, part-of-speech tagging \cite{barrett-etal-2016-pos-tagging}, sentiment analysis \cite{mishra2018cognition,barrett-etal-2018-sequence,long2019improving}, grammatical error detection \cite{barrett-etal-2018-sequence}, hate speech detection \cite{barrett-etal-2018-sequence} and named entity recognition \cite{hollenstein-zhang-2019-entity}.

Different strategies have been adopted to alleviate the need for gaze behaviour at run time. \newcite{barrett-etal-2016-pos-tagging} use token level averages of gaze features at run time from the Dundee Corpus \cite{kennedy2003dundee}, to alleviate the need for gaze behaviour at run time. \newcite{singh-etal-2016-quantifying} and \newcite{long2019improving} predict gaze behaviour at the token-level prior to using it at run time. \newcite{mishra2018cognition}, \newcite{gonzalez2018learning}, \newcite{barrett-etal-2018-sequence}, and \newcite{klerke-etal-2016-improving}, use multi-task learning to learn gaze behaviour along with solving the primary NLP task.

\section{Gaze Behaviour Attributes}
\label{Features Section}

In our experiments, we use only a subset of gaze behaviour attributes described by \newcite{mathias-etal-2018-eyes} because most of the other attributes (like Second Fixation Duration\footnote{The duration of the fixation when the reader fixates on an interest area for the second time.}) were mostly 0, for most of the interest areas, and learning over them would not have yielded any meaningful results.


\paragraph{Fixation Based Attributes} In our experiments, we use the \textbf{Dwell Time} (DT) and \textbf{First Fixation Duration} (FFD) as fixation-based gaze behaviour attributes. Dwell Time is the total amount of time a user spends focusing on an interest area. First Fixation Duration is amount of time that a reader initially focuses on an interest area. Larger values for fixation durations (for both DT and FFD) usually indicate that a word could be wrong (either a spelling mistake or grammar error). Errors would force a reader to pause, as they try to understand why the error was made (For example, if the writer wrote ``short \textbf{\textit{cat}}'' instead of ``short \textbf{\textit{cut}}''.


\paragraph{Saccade Based Attribute} In addition to the Fixation based attributes, we also look at a regression-based attribute - \textbf{IsRegression} (IR). This attribute is used to check whether or not a regression occurred from a given interest area. We don't focus on progression-based attributes, because the usual direction of reading is progressions. We are mainly concerned with regressions because they often occur when there is a mistake, or a need for disambiguation (like trying to resolve the antecedent of an anaphora).


\paragraph{Interest Area Based Attributes} Lastly, we also use IA-based attributes, such as the \textbf{Run Count} (RC) and if the IA was \textbf{Skipped} (Skip). The Run Count is the number of times a particular IA was fixated on, and Skip is whether or not the IA was skipped. A well-written text would be read more easily, meaning a lower RC, and higher Skip \cite{mathias-etal-2018-eyes}.

\section{Dataset and Experiment Setup}
\label{Dataset Section}

\subsection{Essay Dataset Details}

We perform our experiments on the ASAP AEG dataset. The dataset has approximately 13,000 essays, across 8 essay sets. Table \ref{Essay Dataset Details Table} reports the statistics of the dataset in terms of Number of Essays, Score Range, and Mean Word Count. The first 4 rows in Table \ref{Essay Dataset Details Table} are \textbf{\textit{source-dependent response}} (SDR) essay sets, which we use to collect our gaze behaviour data. The other essays are used as \textbf{unseen} essay sets. \textbf{SDRs} are essays written in response to a question about a source article. For example, one of the essay sets that we use is based on an article called \textit{The Mooring Mast}, by Marcia Amidon L{\"u}sted\footnote{The prompt is ``Based on the excerpt, describe the obstacles the builders of the Empire State Building faced in attempting to allow dirigibles to dock there. Support your answer with relevant and specific information from the excerpt.'' The original article is present in Appendix \ref{Source Article Appendix}.}.


\subsection{Evaluation Metric}

\begin{table}[h]
\resizebox{\columnwidth}{!}{
\begin{tabular}{lccc}
\toprule
\textbf{Essay Set} & \textbf{Number of Essays} & \textbf{Score Range} & \textbf{Mean Word Count} \\ \midrule
Prompt 3 & 1726 & 0-3 & 150 \\ 
Prompt 4 & 1770 & 0-3 & 150 \\ 
Prompt 5 & 1805 & 0-4 & 150 \\ 
Prompt 6 & 1800 & 0-4 & 150 \\ \midrule
Prompt 1 & 1783 & 2-12 & 350 \\ 
Prompt 2 & 1800 & 1-6 & 350 \\ 
Prompt 7 & 1569 & 0-30 & 250 \\ 
Prompt 8 & 723 & 0-60 & 650 \\ \midrule
\textbf{Total} & 12976 & 0-60 & 250 \\ \bottomrule
\end{tabular}
}
\caption{Statistics of the 8 essay sets from the ASAP AEG dataset. We collect gaze behaviour data \textbf{\textit{only for Prompts 3 - 6}}, as explained in Section \ref{Gaze Behaviour Dataset Subsection}. The other 4 prompts comprise our \textbf{unseen} essay sets.}
\label{Essay Dataset Details Table}
\end{table}

For measuring our system's performance, we use Cohen's Kappa with quadratic weights - Quadratic Weighted Kappa (QWK) \cite{cohen1968weighted} for the following reasons. Firstly, irrespective of whether we use regression, or ordinal classification, the final scores that are predicted by the system should be discrete scores. Hence, using Pearson Correlation would not be appropriate for our system. Secondly, F-Score and accuracy do not consider chance agreements unlike Cohen's Kappa. If we were to give everyone an average grade, we would get a positive value for accuracy and F-Score, but a Kappa value of 0. Thirdly, \textit{weighted} Kappa takes into account the fact that the classes are ordered, i.e. $0 < 1 < 2...$. Using unweighted Kappa would penalize a $0$ graded as a $4$, as much as a $1$. We use quadratic weights, as opposed to linear weights, because quadratic weights reward agreements and penalize mismatches more than linear weights.

\subsection{Creation of the Gaze Behaviour Dataset}
\label{Gaze Behaviour Dataset Subsection}

In this subsection, we describe how we created our gaze behaviour dataset, how we chose our essays for eye-tracking, and how they were annotated.

\subsubsection{Details of Texts}

\begin{table}[h]
\centering
\begin{tabular}{ccccccc}
\toprule
\textbf{Essay Set} & \textbf{0} & \textbf{1} & \textbf{2} & \textbf{3} & \textbf{4} & \textbf{Total} \\ \midrule
Prompt 3 & 2 & 4 & 5 & 1 & N/A & 12 \\ 
Prompt 4 & 2 & 3 & 4 & 3 & N/A & 12 \\ 
Prompt 5 & 2 & 1 & 3 & 5 & 1 & 12 \\ 
Prompt 6 & 2 & 2 & 3 & 4 & 1 & 12 \\ \midrule
\textbf{Total} & 8 & 10 & 15 & 13 & 2 & 48 \\ \bottomrule
\end{tabular}%
\caption{Number of essays for each essay set which we collected gaze behaviour, scored between 0 to 3 (or 4).}
\label{Gaze Dataset Details Table}
\end{table}


As mentioned earlier in Section \ref{Dataset Section}, we used only essays corresponding to prompts 3 to 6 of the ASAP AEG dataset. From \textbf{each of the four essay sets}, we selected \textbf{12 essays} with a diverse vocabulary as well as all possible scores.

We use a greedy algorithm to select essays \textit{i.e.,} For each essay set, we pick 12 essays, covering all score points with maximum number of unique tokens, as well as being under 250 words. Table \ref{Gaze Dataset Details Table} reports the distribution of essays with each score, for each of the 4 essay sets that we use to create our gaze behaviour dataset.

To display the essay text on the screen, we use a large font size, so that (a) the text is clear, and (b) the reader's gaze is captured on the words which they are currently reading. Although, this ensures the clarity in reading and recording the gaze pattern in a more accurate manner, it also imposes a limitation on the size of the essay which can be used for our experiment. This is why, the longest essay in our gaze behaviour dataset is \textbf{about 250 words}.

The original essays have their named entities anonymized. Hence, before running the experiments, we replaced the required named entities with placeholders (Eg. @NAME1 $\rightarrow$ ``Al Smith'', @PLACE1 $\rightarrow$ ``New Jersey'', @MONTH1 $\rightarrow$ ``May'', etc.)\footnote{Another advantage of using source-dependent essays is that there is a source article which we can use to correctly replace the anonymized named entities}.

\subsubsection{Annotator Details}


We used a total of \textbf{8 annotators}, \textbf{aged between 18 and 31}, with an \textbf{average age of 25 years}. All of them were either in college, or had completed a Bachelor's degree. All but one of them also had experience as a teaching assistant. The annotators were fluent in English, and about half of them had participated earlier, in similar experiments. The annotators were adequately compensated for their work\footnote{We report details on individual annotators in Appendix \ref{Annotator Profiles Appendix}.}.


To assess the quality of the individual annotators, we evaluated the scores they provided against the ground truth scores - \textit{i.e.,} the scores given by the original annotators. The \textbf{QWK} measures the agreement between the annotators and the ground truth score. \textbf{Close} is the number of times (out of 48) in which the annotators either agreed with the ground truth scores, or differed from them by \textbf{at most 1} score point. \textbf{Correct} is the number of times (out of 48) in which the annotators agreed with the ground truth scores. The mean values for the 3 measures were \textbf{0.646} (QWK), \textbf{42.75} (Close) and \textbf{22.25} (Correct).

\subsection{System Details}
\label{System Details Subsection}
We conduct our experiments using well-established norms in eye-tracking research \cite{holmqvist2011eye}. The essays are displayed on a screen that is kept about \textbf{2 feet} in front of the participant.


The workflow of the experiment is as follows. First, \textbf{the camera is calibrated}. This is done by having the annotator look at \textbf{13 points} on the screen, while the camera tracks their eyes. Next, \textbf{the calibration is validated}. In this step, the participant looks at the same points they saw earlier. If there is a big difference between the participant's fixation points tracked by the camera and the actual points, calibration is repeated. Then, \textbf{the reader performs a self-paced reading of the essay} while we supervise the tracking of their eyes. After reading and scoring an essay, the participant takes a small break of \textbf{about a minute}, before continuing. Before the next essay is read, the camera has to again be calibrated and validated\footnote{The average time for the participants was about 2 hours, with the fastest completing the task in slightly under one and a half hours.}. The essay is displayed on the screen in \textbf{Times New Roman} typeface with a \textbf{font size of 23}. Finally, \textbf{the reader scores the essay} and \textbf{provides a justification for their score}\footnote{As part of our data release, we will release the scores given by each annotator, as well as their justifications for their score}.





This entire process is done using an \textbf{SR Research Eye Link 1000} eye-tracker (monocular stabilized head mode, with a sampling rate of 500Hz). The machine collects all the gaze details that we need for our experiments. An interest area report is generated for gaze behaviour using the \textbf{SR Research Data Viewer} software.

\begin{figure*}[t]
\centering
\includegraphics[width=\textwidth]{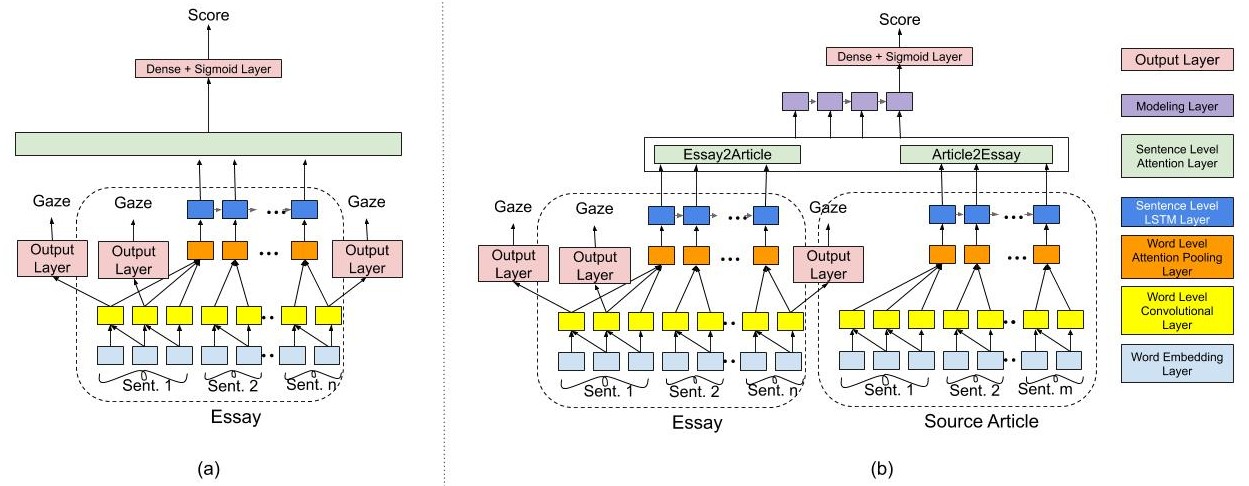}
\caption{Architecture of the proposed gaze behaviour and essay scoring multi-task learning systems, namely (a) - the \textbf{Self-Attention} multi-task learning system, for an essay of $n$ sentences - and (b) - the \textbf{Co-Attention} system for an essay of $n$ sentences and a source article of $m$ sentences.}
\label{fig:muliTaskAES}
\end{figure*}

\subsection{Experiment Details}

We use \textbf{five-fold cross-validation} to evaluate our system. For each fold, \textbf{60\%} is used as training, \textbf{20\%} for validation, and \textbf{20\%} for testing. The folds are the same as those used by \newcite{taghipour-ng-2016-neural}. Prior to running our experiments, we convert the scores from their original score range (given in Table \ref{Essay Dataset Details Table}) to the \textbf{range of} $[0 , 1]$ as described by \newcite{taghipour-ng-2016-neural}.


In order to normalize idiosyncratic reading patterns across different readers, we perform binning for each of the features for each of the readers. For IR and Skip we use only two bins - 0 and 1 - corresponding to their values. For the run count, we use six bins (from 0 to 5), where each bin is the run count (up to 4), and bin 5 contains run counts \textbf{\textit{more than 4}}. For the fixation attributes - DT and FFD - we use the same binning scheme as described in \newcite{klerke-etal-2016-improving}. The binning scheme for fixation attributes is as follows:

$0$ if $FV = 0$,

$1$ if $FV > 0$ and $FV \leq \mu - \sigma$,

$2$ if $FV > \mu - \sigma$ and $FV \leq \mu - 0.5 \times \sigma$,

$3$ if $FV > \mu - 0.5 \times \sigma $ and $FV \leq \mu + 0.5 \times \sigma$,

$4$ if $FV > \mu + 0.5 \times \sigma$ and $FV \leq \mu + \sigma$,

$5$ if $FV > \mu + \sigma$,

\noindent where $FV$ is the value of the given fixation attribute, $\mu$ is the average fixation attribute value for the reader and $\sigma$ is the standard deviation.

\subsection{Network Architecture}


Figure \ref{fig:muliTaskAES} (b) shows the architecture of our proposed system, based on the co-attention based architecture described by \newcite{zhang-litman-2018-co}. Given an essay, we split the essay into sentences. For each sentence, we look-up the word embeddings for all words in the \textbf{Word Embedding} layer. The \textbf{4000 most frequent words} are used as the vocabulary, with all other words mapped to a special unknown token. This sequence of word embeddings is then sent through a Time-Delay Neural Network (TDNN), or \textbf{1-d Convolutional Neural Network} (CNN), of filter width $k$. The output from CNN is pooled using an attention layer - the \textbf{Word Level Attention Pooling Layer} - which results in a representation for every sentence. These sentence representations are then sent through a \textbf{Sentence Level LSTM Layer} and their output pooled in the \textbf{Sentence Level Attention Pooling Layer} to obtain the \textit{sentence representation for the essay}.

A similar procedure is repeated for the source article. We then perform co-attention between the sentence representations of \textbf{the essay} and \textit{the source article}. \textbf{Co-attention} is performed to learn similarities between the sentences in the essay and the source article. This is done as a way to ensure that the writer sticks to answering the prompt, rather than drifting off topic. 

We now represent every sentence in the essay as a weighted combination of the sentence representation between the essay and the source article (Essay2Article). The weights are obtained from the output of the co-attention layer. The weights represent how each sentence in the essay are similar to the sentences in the source article. If a sentence in the essay has low weights this indicates that the sentence would be off topic. A similar procedure is repeated to get a weighted representation of sentences in the source article with respect to the essay  (Article2Essay).

Finally, we send the sentence representation of the essay and article, through a dense layer (i.e. the \textbf{Modeling Layer}) to predict the final essay score, with a \textbf{sigmoid activation function}. As the essay scores are in the range $[0,1]$, we use sigmoid activation at the output layer. During prediction, we map the output scores from the sigmoid layer back to the original score range, minimizing the \textbf{mean squared error (MSE) loss}.


For essay sets without a source article, we use the \textit{Self-Attention} model proposed by \newcite{dong-etal-2017-attention}. This is a simpler model which does not consider the source article, and uses only the essay text. This is applicable whenever a source article is not present. Figure \ref{fig:muliTaskAES} (a) shows the architecture of the model. Like the earlier system, we get the \textit{sentence representation of the essay} from the \textbf{Sentence Level LSTM Layer} and send it through the Dense Layer with a sigmoid activation function.

Gaze behaviour is learnt at the Word-Level Convolutional Layer in both the models because the gaze attributes are defined at the word-level, while the essay is scored at the document-level. The output from the CNN layer is sent through a linear layer followed by sigmoid activation for a particular gaze behaviour. For learning multiple gaze attributes simultaneously, we have multiple linear layers for each of the gaze attributes. In the multi-task setting, we also minimize the mean squared error of the learnt gaze behaviour and the actual gaze behaviour attribute value. We assign weights to each of the gaze behaviour loss functions to control the importance given to individual gaze behaviour learning tasks.

\subsection{Network Hyperparameters}
\label{Network Hyperparameters Subsection}

\begin{table}[h]
\resizebox{\columnwidth}{!}{
\begin{tabular}{llc}
\toprule
\textbf{Layer} & \textbf{Hyperparameter} & \textbf{Value}\\ \midrule
Embedding layer & Pre-trained embeddings & GloVe \\ 
 & Embeddings dimensions & 50 \\ \midrule
Word-level CNN & Kernel size & 5 \\
 & Filters & 100 \\ \midrule
Sentence-level LSTM & Hidden units & 100 \\ \midrule
Network-wide & Batch size & 100 \\
 & Epochs & 100 \\
 & Learning rate & 0.001 \\
 & Dropout rate & 0.5 \\
 & Momentum & 0.9 \\
\bottomrule
\end{tabular}
}
\caption{Hyperparameters for our experiment.}
\label{Hyperparameters Table}
\end{table}

Table \ref{Hyperparameters Table} gives the different hyperparameters which we used in our experiment. We use the 50 dimension GloVe pre-trained word embeddings \cite{pennington-etal-2014-glove} trained on the Wikipedia 2014 + Gigawords 5 Corpus (6B tokens, 4K vocabulary, uncased). We run our experiments over a batch size of 100, for 100 epochs, and set the learning rate as 0.001, and a dropout rate of 0.5. The Word-level CNN layer has a kernel size of 5, with 100 filters. The Sentence-level LSTM layer and modeling layer both have 100 hidden units. We use the RMSProp Optimizer \cite{dauphin2015rmsprop} with a 0.001 initial learning rate and momentum of 0.9. 



\begin{table}[h]
\centering
\begin{tabular}{lc}
\toprule
\textbf{Gaze Feature} & \textbf{Gaze Feature Weight} \\ \bottomrule
Dwell Time & 0.05 \\
First Fixation Duration & 0.05 \\ \midrule
IsRegression & 0.01 \\ \midrule
Run Count & 0.01 \\
Skip & 0.1 \\ \bottomrule
\end{tabular}%
\caption{This table shows the \textbf{best weights} assigned to the different gaze features from our grid search.}
\label{Weighting Table}
\end{table}

In addition to the network hyper-parameters, we also weigh the loss functions of the different gaze behaviours differently, with weight levels of \textbf{0.5}, \textbf{0.1}, \textbf{0.05}, \textbf{0.01} and \textbf{0.001}. We use grid search and pick the weight giving the lowest mean-squared error on the \textit{development} set. The best weights from grid search are \textbf{0.05} for DT and FFD, \textbf{0.01} for IR and RC, and \textbf{0.1} for Skip. 


\begin{table*}[t]
\centering
\resizebox{0.9\textwidth}{!}{%
\begin{tabular}{lccccc}
\toprule
\textbf{System} & \textbf{Prompt 3} & \textbf{Prompt 4} & \textbf{Prompt 5} & \textbf{Prompt 6} & \textbf{Mean QWK} \\ \midrule
\newcite{taghipour-ng-2016-neural} & 0.683 & 0.795 & \textbf{0.818} & 0.813 & 0.777 \\ 
\newcite{dong-zhang-2016-automatic} & 0.662 & 0.778 & 0.800 & 0.809 & 0.762 \\ 
\newcite{tay-2018-skipflow} & 0.695 & 0.788 & 0.815 & 0.810 & 0.777 \\ 
\midrule
Self-Attention \cite{dong-etal-2017-attention} & 0.677 & 0.807 & 0.806 & 0.809 & 0.775 \\ 
Co-Attention \cite{zhang-litman-2018-co} & 0.689$\dagger$ & 0.809$\dagger$ & 0.812$\dagger$ & 0.813$\dagger$ & 0.780$\dagger$ \\ \midrule
Co-Attention+Gaze & \textbf{0.698*} & \textbf{0.818*} & 0.815* & \textbf{0.821*} & \textbf{0.788*} \\ 
\bottomrule
\end{tabular}%
}
\caption{Results of our experiments in scoring the essays (QWK values) from the essay sets where we collected gaze behaviour. The first 3 rows are results reported from other state-of-the-art deep learning systems. The next 2 rows are the results we obtained on existing systems - self-attention and co-attention - without gaze behaviour. The last row is the results from our system using gaze behaviour data (Co-Attention+Gaze). $\dagger$ denotes the baseline system performance, and * denotes a statistically significant result of $p<0.05$ for the gaze behaviour system.}
\label{Results Table}
\end{table*}

\begin{table*}[t]
\centering
\resizebox{0.9\textwidth}{!}{%
\begin{tabular}{lccccc}
\toprule
\textbf{System} & \textbf{Prompt 1} & \textbf{Prompt 2} & \textbf{Prompt 7} & \textbf{Prompt 8} & \textbf{Mean QWK} \\ \midrule
\newcite{taghipour-ng-2016-neural} & 0.775 & \textbf{0.687} & 0.805 & 0.594 & 0.715 \\ 
\newcite{dong-zhang-2016-automatic} & 0.805 & 0.613 & 0.758 & 0.644 & 0.705 \\ 
\newcite{tay-2018-skipflow} & 0.832 & 0.684 & 0.800 & 0.697 & 0.753 \\ 
\midrule
Only Prompt (\newcite{dong-etal-2017-attention}) & 0.816 & 0.667 & 0.792 & 0.678 & 0.738 \\ 
Extra Essays & 0.828$\dagger$ & 0.672$\dagger$ & 0.802$\dagger$ & 0.685$\dagger$ & 0.747$\dagger$ \\ \midrule
Extra Essays + Gaze & \textbf{0.833} & 0.681 & \textbf{0.806*} & \textbf{0.699*} & \textbf{0.754*} \\ 
\bottomrule
\end{tabular}%
}
\caption{Results of our experiments on the \textbf{unseen} essay sets our dataset. The first 3 rows are results reported from other state-of-the-art deep learning systems. The next 2 rows are the results obtained without using gaze behaviour (without and with the extra essays). The last row is the results from our system. $\dagger$ denotes the baseline system without gaze behaviour, and * denotes a statistically significant result of $p<0.05$ for the gaze behaviour system.}
\label{Unseen Prompts Result Table}
\end{table*}

\subsection{Experiment Configurations}

To test our system on essay sets which we collected gaze behaviour, we run experiments using the following configurations. (a) \textbf{Self-Attention} -  This is the implementation of \newcite{dong-etal-2017-attention}'s system in Tensorflow by \newcite{zhang-litman-2018-co}. (b) \textbf{Co-Attention}. This is \newcite{zhang-litman-2018-co}'s system\footnote{The implementation of both systems can be downloaded from \href{https://github.com/Rokeer/co-attention}{here}.}. (c) \textbf{Co-Attention+Gaze}. This is our system, which uses gaze behaviour.


In addition to this, we also run experiments on the \textbf{unseen} essay sets using the following \textit{training} configurations. (a) \textbf{Only Prompt} - This uses our self-attention model, with the training data being only the essays from that essay set. We use this model, because there are no source articles for these essay sets. (b) \textbf{Extra Essays} - Here, we augment the training data of (a) with the \textbf{\textit{48 essays}} for which we collect gaze behaviour data. (c) \textbf{Essays+Gaze} - Here, we augment the training data of (a) with the \textbf{\textit{48 essays}} which we collect gaze behaviour data, and their corresponding \textbf{\textit{gaze data}}. We also compare our results with a string kernel based system proposed by \newcite{cozma-etal-2018-automated}.


\begin{figure*}[t]
\centering
\fbox{\includegraphics[width=0.85\textwidth]{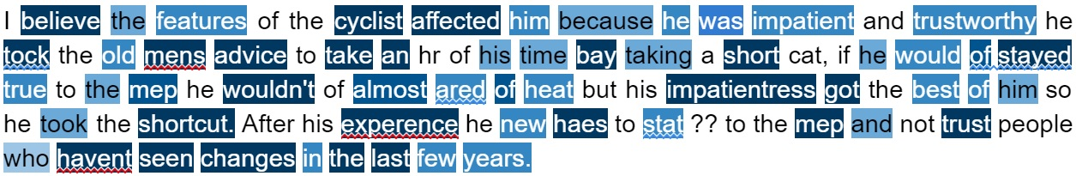}}
\caption{\textbf{Dwell Time} of one of the readers for one of the essays. The darker the background, the larger the bin.} 
\label{Dwell Time Heat Map Image}
\end{figure*}





\section{Results and Analysis}
\label{Results Section}

Table \ref{Results Table} reports the results of our experiments on the essay sets for which we collect the gaze behaviour data. The table is divided into 3 parts. The first part (\textit{i.e.,} first 3 rows) are the reported results previously available deep-learning systems, namely \newcite{taghipour-ng-2016-neural}, \newcite{dong-zhang-2016-automatic}, and \newcite{tay-2018-skipflow}. The next 2 rows feature results using the self-attention \cite{dong-etal-2017-attention} and co-attention \cite{zhang-litman-2018-co}. The last row reports results using gaze behaviour on top of co-attention, \textit{i.e.,} Co-Attention+Gaze. The first column is the different systems. The next 4 columns report the QWK results of each system for each of the 4 essay sets. The last column reports the Mean QWK value across all 4 essay sets. 

Our system is able to outperform the Co-Attention system \cite{zhang-litman-2018-co} in all the essay sets. Overall, it is also the best system - achieving the highest QWK results among all the systems in 3 out of the 4 essay sets (and the second-best in the other essay set). To test our hypothesis - that the model trained by learning gaze behaviour helps in automatic essay grading - we run the Paired T-Test. Our null hypothesis is: ``Learning gaze behaviour to score an essay does not help any more than the self-attention and co-attention systems and whatever improvements we see are due to chance.'' We choose a significance level of $p < 0.05$, and observe that the improvements of our system are found to be statistically significant - rejecting the null hypothesis.


\subsection{Results for Unseen Essay Sets}

In order to run our experiments on \textbf{unseen} essay sets, we augment the training data with the gaze behaviour data collected. Since none of these essays have source articles, we use the self-attention model of \newcite{dong-etal-2017-attention} as the baseline system. We now augment the gaze behaviour learning task as the auxiliary task and report the results in Table \ref{Unseen Prompts Result Table}. The first column in the table is the different systems. The next 4 columns are the results for each of the unseen essay sets, and the last column is the mean QWK. From Table \ref{Unseen Prompts Result Table}, we observe that our system which uses both the \textbf{extra 48 essays and their gaze behaviour} outperforms the other 2 configurations (\textbf{Only Prompt} and \textbf{Extra Essays}) across all 4 unseen essay sets. The improvement when learning gaze behaviour for \textbf{unseen} essay sets is statistically significant for $p<0.05$.
\subsection{Comparison with String Kernel System}

Since \citet{cozma-etal-2018-automated} haven't released their data splits (train/test/dev), we ran their system with our data splits. We observed a mean QWK of \textbf{0.750} with the string kernel-based system on the essay sets where we have gaze behaviour data, and \textbf{0.685} on the unseen essay sets. One possible reason for this could be that while they used cross-validation, they may have used only a training-testing split (as compared to a train/test/dev split).

\subsection{Analysis of Gaze Attributes}



\begin{table}[h]
\centering
\begin{tabular}{lc}
\toprule
\textbf{Gaze Feature} & \textbf{Diff. in QWK} \\ \bottomrule
Dwell Time & \textbf{0.0137} \\
First Fixation Duration & 0.0136 \\ \midrule
IsRegression & 0.0090 \\ \midrule
Run Count & 0.0110 \\
Skip & 0.0091 \\ \bottomrule
\end{tabular}%
\caption{Results of ablation tests for each gaze behaviour attribute across all the essay sets. The reported numbers are the \textbf{difference in QWK} before and after ablating the given gaze attribute. The number in \textbf{bold} denotes the best gaze attribute.}
\label{Gaze Feature Ablation Test Table}
\end{table}

In order to see which of the gaze attributes are the most important, we ran ablation tests, where we ablate each gaze attribute.
We found that the most important gaze behaviour attribute across all the essay sets is the Dwell Time, followed closely by the First Fixation Duration. One of the reasons for this is the fact that both DT and FFD were very useful in detecting errors made by the essay writers. From Figure \ref{Dwell Time Heat Map Image}\footnote{We have given more examples in Appendix \ref{Heat Maps Appendix}.}, we observe that most of the longest dwell times have come at/around spelling mistakes (\textbf{\textit{tock}} instead of \textbf{\textit{took}}), or out-of-context words (\textbf{\textit{bay}} instead of \textbf{\textit{by}}), or incorrect phrases (\textbf{\textit{short cat}}, instead of \textbf{\textit{short cut}}). These errors force the reader to spend more time fixating on the word which we also mentioned earlier.

The \textbf{normalized MSE} of each of the gaze features learnt by our system was between \textbf{0.125 to 0.128} for all the gaze behaviour attributes.

\subsection{Analysis Using Only a Native English Speaker}

\begin{table}[h]
\centering
\begin{tabular}{lccc}
\toprule
\textbf{System} & \textbf{No} & \textbf{Native} & \textbf{All} \\ \midrule
Prompt 1 & 0.816 & 0.824 & 0.833 \\
Prompt 2 & 0.667 & 0.679 & 0.681 \\
Prompt 3 & 0.677 & 0.679 & 0.698 \\
Prompt 4 & 0.807 & 0.812 & 0.818 \\
Prompt 5 & 0.806 & 0.810 & 0.815 \\
Prompt 6 & 0.809 & 0.815 & 0.821 \\
Prompt 7 & 0.792 & 0.809 & 0.806 \\
Prompt 8 & 0.678 & 0.679 & 0.699 \\ \midrule
\textbf{Mean QWK} & 0.757 & 0.764 & 0.771 \\
\bottomrule
\end{tabular}
\caption{Result using only gaze behaviour of the native speaker (Native), compared using no gaze behaviour (No) and gaze behaviour of all the readers (All).}
\label{Native Results Table}
\end{table}

We also ran our experiments using only the gaze behaviour of an annotator who was a \textbf{native} English speaker (as opposed to the rest of our annotators who were just \textit{fluent} English speakers). Table \ref{Native Results Table} shows the results of those experiments. We observed a mean QWK of \textbf{0.779} for the seen essay sets, and a mean QWK of \textbf{0.748} for the essays sets where we have no gaze data. The difference in performance between both our systems (i.e. with only native speaker and with all annotators) were found to be statistically significant with $p=0.0245$\footnote{The p-values for the different experiments are in Appendix \ref{P-Values Appendix}.}. Similarly, the improvement in performance using the native English speaker, compared to not using any gaze behaviour was also found to be statistically significant for $p=0.0084$.



\section{Conclusion and Future Work}
\label{Conclusion Section}

In this paper, we describe how learning gaze behaviour can help AEG in a multi-task learning setup. We explained how we created a resource by collecting gaze behaviour data, and using multi-task learning we are able to achieve better results over a state-of-the-art system developed by \newcite{zhang-litman-2018-co} for the essay sets which we collected gaze behaviour data from. We also analyze the transferability of gaze behaviour patterns across essay sets by training a multi-task learning model on \textbf{unseen} essay sets (i.e. essay sets where we have no gaze behaviour data), thereby establishing that learning gaze behaviour  improves automatic essay grading.


In the future, we would like to look at using gaze behaviour to help in cross-domain AEG. This is done mainly when we don't have enough training examples in our essay set. We would also like to explore the possibility of generating textual feedback (rather than just a number, denoting the score of the essay) based on the justifications that the annotators gave for their grades.


\bibliography{anthology,aacl-ijcnlp2020}
\bibliographystyle{acl_natbib}
\appendix
\section{Source Article (Prompt 6)}
\label{Source Article Appendix}

\textit{The Mooring Mast}, by Marcia Amidon L\"{u}sted

When the Empire State Building was conceived, it was planned as the world’s tallest building, taller even than the new Chrysler Building that was being constructed at Forty-second Street and Lexington Avenue in New York. At seventy-seven stories, it was the tallest building before the Empire State began construction, and Al Smith was determined to outstrip it in height.

The architect building the Chrysler Building, however, had a trick up his sleeve. He secretly constructed a 185-foot spire inside the building, and then shocked the public and the media by hoisting it up to the top of the Chrysler Building, bringing it to a height of 1,046 feet, 46 feet taller than the originally announced height of the Empire State Building.

Al Smith realized that he was close to losing the title of world’s tallest building, and on December 11, 1929, he announced that the Empire State would now reach the height of 1,250 feet. He would add a top or a hat to the building that would be even more distinctive than any other building in the city. John Tauranac describes the plan:

``[The top of the Empire State Building] would be more than ornamental, more than a spire or dome or a pyramid put there to add a desired few feet to the height of the building or to mask something as mundane as a water tank. Their top, they said, would serve a higher calling. The Empire State Building would be equipped for an age of transportation that was then only the dream of aviation pioneers.''

This dream of the aviation pioneers was travel by dirigible, or zeppelin, and the Empire State Building was going to have a mooring mast at its top for docking these new airships, which would accommodate passengers on already existing transatlantic routes and new routes that were yet to come.

\subsection{The Age of Dirigibles}

By the 1920s, dirigibles were being hailed as the transportation of the future. Also known today as blimps, dirigibles were actually enormous steel-framed balloons, with envelopes of cotton fabric filled with hydrogen and helium to make them lighter than air. Unlike a balloon, a dirigible could be maneuvered by the use of propellers and rudders, and passengers could ride in the gondola, or enclosed compartment, under the balloon.

Dirigibles had a top speed of eighty miles per hour, and they could cruise at seventy miles per hour for thousands of miles without needing refueling. Some were as long as one thousand feet, the same length as four blocks in New York City. The one obstacle to their expanded use in New York City was the lack of a suitable landing area. Al Smith saw an opportunity for his Empire State Building: A mooring mast added to the top of the building would allow dirigibles to anchor there for several hours for refueling or service, and to let passengers off and on. Dirigibles were docked by means of an electric winch, which hauled in a line from the front of the ship and then tied it to a mast. The body of the dirigible could swing in the breeze, and yet passengers could safely get on and off the dirigible by walking down a gangplank to an open observation platform.

The architects and engineers of the Empire State Building consulted with experts, taking tours of the equipment and mooring operations at the U.S. Naval Air Station in Lakehurst, New Jersey. The navy was the leader in the research and development of dirigibles in the United States. The navy even offered its dirigible, the Los Angeles, to be used in testing the mast. The architects also met with the president of a recently formed airship transport company that planned to offer dirigible service across the Pacific Ocean.

When asked about the mooring mast, Al Smith commented:

``[It’s] on the level, all right. No kidding. We’re working on the thing now. One set of engineers here in New York is trying to dope out a practical, workable arrangement and the Government people in Washington are figuring on some safe way of mooring airships to this mast.''

\subsection{Designing the Mast}

The architects could not simply drop a mooring mast on top of the Empire State Building’s flat roof. A thousand-foot dirigible moored at the top of the building, held by a single cable tether, would add stress to the building’s frame. The stress of the dirigible’s load and the wind pressure would have to be transmitted all the way to the building’s foundation, which was nearly eleven hundred feet below. The steel frame of the Empire State Building would have to be modified and strengthened to accommodate this new situation. Over sixty thousand dollars’ worth of modifications had to be made to the building’s framework.

Rather than building a utilitarian mast without any ornamentation, the architects designed a shiny glass and chrome-nickel stainless steel tower that would be illuminated from inside, with a stepped-back design that imitated the overall shape of the building itself. The rocket-shaped mast would have four wings at its corners, of shiny aluminum, and would rise to a conical roof that would house the mooring arm. The winches and control machinery for the dirigible mooring would be housed in the base of the shaft itself, which also housed elevators and stairs to bring passengers down to the eighty-sixth floor, where baggage and ticket areas would be located.

The building would now be 102 floors, with a glassed-in observation area on the 101st floor and an open observation platform on the 102nd floor. This observation area was to double as the boarding area for dirigible passengers.

Once the architects had designed the mooring mast and made changes to the existing plans for the building’s skeleton, construction proceeded as planned. When the building had been framed to the 85th floor, the roof had to be completed before the framing for the mooring mast could take place. The mast also had a skeleton of steel and was clad in stainless steel with glass windows. Two months after the workers celebrated framing the entire building, they were back to raise an American flag again—this time at the top of the frame for the mooring mast.

\subsection{The Fate of the Mast}

The mooring mast of the Empire State Building was destined to never fulfill its purpose, for reasons that should have been apparent before it was ever constructed. The greatest reason was one of safety: Most dirigibles from outside of the United States used hydrogen rather than helium, and hydrogen is highly flammable. When the German dirigible Hindenburg was destroyed by fire in Lakehurst, New Jersey, on May 6, 1937, the owners of the Empire State Building realized how much worse that accident could have been if it had taken place above a densely populated area such as downtown New York.

The greatest obstacle to the successful use of the mooring mast was nature itself. The winds on top of the building were constantly shifting due to violent air currents. Even if the dirigible were tethered to the mooring mast, the back of the ship would swivel around and around the mooring mast. Dirigibles moored in open landing fields could be weighted down in the back with lead weights, but using these at the Empire State Building, where they would be dangling high above pedestrians on the street, was neither practical nor safe.

The other practical reason why dirigibles could not moor at the Empire State Building was an existing law against airships flying too low over urban areas. This law would make it illegal for a ship to ever tie up to the building or even approach the area, although two dirigibles did attempt to reach the building before the entire idea was dropped. In December 1930, the U.S. Navy dirigible Los Angeles approached the mooring mast but could not get close enough to tie up because of forceful winds. Fearing that the wind would blow the dirigible onto the sharp spires of other buildings in the area, which would puncture the dirigible’s shell, the captain could not even take his hands off the control levers.

Two weeks later, another dirigible, the Goodyear blimp Columbia, attempted a publicity stunt where it would tie up and deliver a bundle of newspapers to the Empire State Building. Because the complete dirigible mooring equipment had never been installed, a worker atop the mooring mast would have to catch the bundle of papers on a rope dangling from the blimp. The papers were delivered in this fashion, but after this stunt the idea of using the mooring mast was shelved. In February 1931, Irving Clavan of the building’s architectural office said, ``The as yet unsolved problems of mooring air ships to a fixed mast at such a height made it desirable to postpone to a later date the final installation of the landing gear.''

By the late 1930s, the idea of using the mooring mast for dirigibles and their passengers had quietly disappeared. Dirigibles, instead of becoming the transportation of the future, had given way to airplanes. The rooms in the Empire State Building that had been set aside for the ticketing and baggage of dirigible passengers were made over into the world’s highest soda fountain and tea garden for use by the sightseers who flocked to the observation decks. The highest open observation deck, intended for disembarking passengers, has never been open to the public.

\begin{table*}[t]
\centering
\resizebox{\textwidth}{!}{
\begin{tabular}{|l|l|c|l|c|c|c|c|c|c|}
\hline
\textbf{ID} & \textbf{Sex} & \textbf{Age} &  \textbf{Occupation} & \textbf{TA?} & \textbf{L1 Language} & \textbf{English Score} & \textbf{QWK} & \textbf{Correct} & \textbf{Close} \\
\hline
Annotator 1 & Male & 23 & Masters student & Yes & Hindi & 94\% & 0.611 & 19 & 41 \\
Annotator 2 & Male & 18 & Undergraduate & Yes & Marathi & 95\% & 0.587 & 24 & 41 \\
Annotator 3 & Male & 31 & Research scholar & Yes & Marathi & 85\% & 0.659 & 21 & 43 \\
Annotator 4 & Male & 28 & Software engineer & Yes & English & 96\% & 0.659 & 26 & 44 \\
Annotator 5 & Male & 30 & Research scholar & Yes & Gujarati & 92\% & 0.600 & 19 & 42 \\
Annotator 6 & Female & 22 & Masters student & Yes & Marathi & 95\% & 0.548 & 19 & 40 \\
Annotator 7 & Male & 19 & Undergraduate & Yes & Marathi & 93\% & 0.732 & 21 & 46 \\
Annotator 8 & Male & 28 & Masters student & Yes & Gujarati & 94\% & 0.768 & 29 & 45 \\
\hline
\end{tabular}
}
\caption{Profile of the annotators}
\label{Annotator Table}
\end{table*}

\section{Annotator Profiles}
\label{Annotator Profiles Appendix}

Table \ref{Annotator Table} summarizes the profiles of the different annotators. It details each of the 8 annotators, their sex, age, occupations, L1 / native languages, their performance in a high school Examination in English and whether or not they have had experience as a TA. The last 3 columns are their performance on the annotation grading task, where QWK is their agreement with the ground truth scores, Correct is the number of times (out of 48) where their essay scores matched with the ground truth scores, and Close is the number of times (out of 48) where they disagreed with the ground truth score by at most 1 grade point.

\section{Heat Map Examples}
\label{Heat Maps Appendix}

\subsection{Different Gaze Features}

Here, we show examples of heat maps for different gaze behaviour attributes of one of our readers.

\begin{figure*}[t]
\centering
\includegraphics[width=\textwidth]{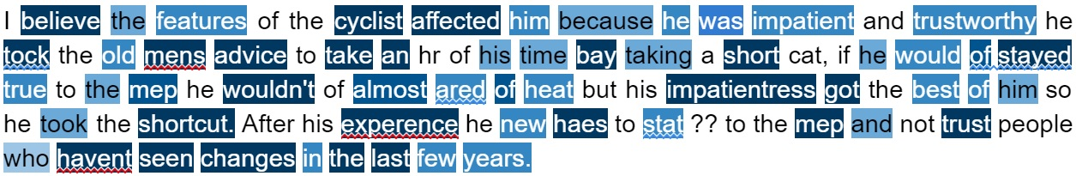}
\caption{Sample heat map of the dwell of a reader for the text. The darker the blue, the larger the bin, and the longer the dwell time.}
\label{Dwell Heat Map Example Figure}
\end{figure*}

\begin{figure*}[t]
\centering
\includegraphics[width=\textwidth]{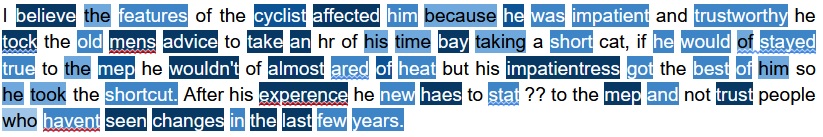}
\caption{Sample heat map of the first fixation duration of a reader for the text. The darker the blue, the larger the bin, and the longer the first fixation duration.}
\label{First Fixation Duration Heat Map Example Figure}
\end{figure*}

\begin{figure*}[t]
\centering
\includegraphics[width=\textwidth]{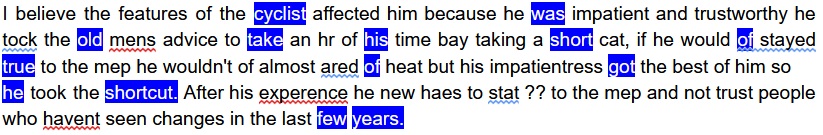}
\caption{Sample heat map of the Is Regression feature of a reader for the text. The highlighted words denote words that the reader regressed from.}
\label{Is Regression Heat Map Example Figure}
\end{figure*}

\begin{figure*}[t]
\centering
\includegraphics[width=\textwidth]{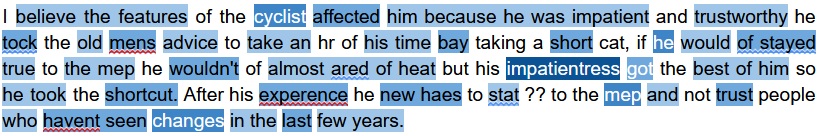}
\caption{Sample heat map of the run count of a reader for the text. The darker the blue, the larger the bin, and the higher the run count.}
\label{Run Count Heat Map Example Figure}
\end{figure*}

\begin{figure*}[t]
\centering
\includegraphics[width=\textwidth]{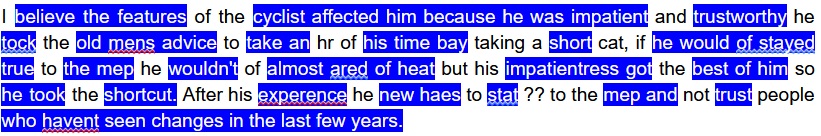}
\caption{Sample heat map of the Skip feature of a reader for the text. The \textbf{unhighlighted} words denote words that the reader skipped.}
\label{Skip Heat Map Example Figure}
\end{figure*}

\begin{enumerate}
\item Figure \ref{Dwell Heat Map Example Figure} shows the dwell time of the
reader.
\item Figure \ref{First Fixation Duration Heat Map Example Figure} shows the heat map of the first fixation duration of a reader.
\item Figure \ref{Is Regression Heat Map Example Figure} shows the heat map of the IsRegression feature - i.e. whether or not the reader regressed from a particular word.
\item Figure \ref{Run Count Heat Map Example Figure} shows the heat map of the Run Count of the reader.
\item Figure \ref{Skip Heat Map Example Figure} shows the words that the reader read (highlighted) and skipped (unhighlighted).
\end{enumerate}

\subsection{Dwell Times of Good and Bad Essays}

\begin{figure*}[t]
\centering
\includegraphics[width=\textwidth]{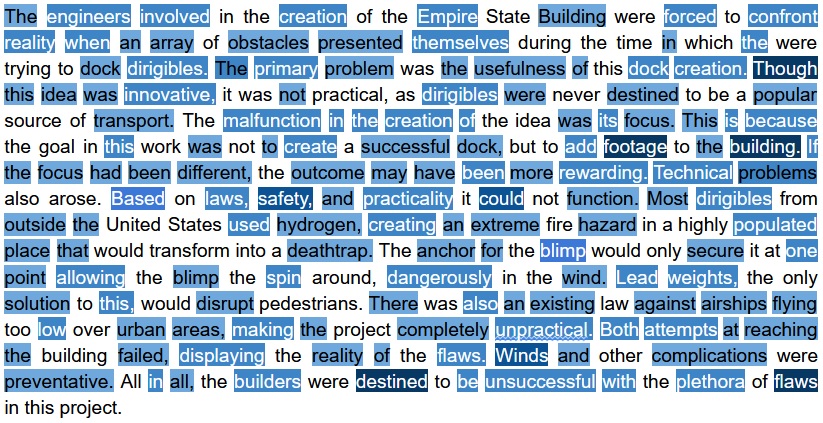}
\caption{Dwell Time for a reader for an essay which he scored well.}
\label{Good Essay Example}
\end{figure*}

\begin{figure*}[t]
\centering
\includegraphics[width=\textwidth]{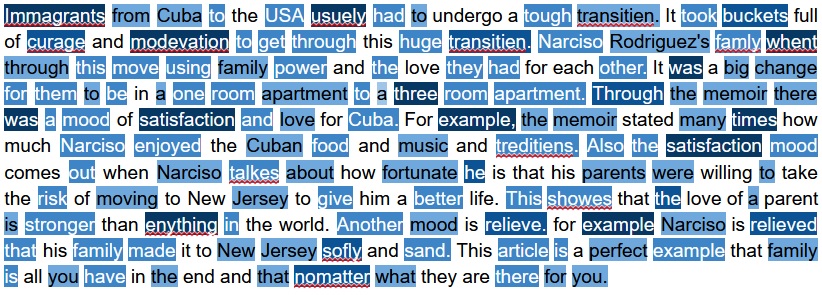}
\caption{Dwell Time for a reader for an essay which he scored badly.}
\label{Bad Essay Example}
\end{figure*}

Figures \ref{Good Essay Example} and \ref{Bad Essay Example} show the dwell time heat maps of a reader as he reads a good essay and a bad essay respectively. For the bad essay, notice the amount of a lot more darker blues compared to the good essay.

\section{P-Values}
\label{P-Values Appendix}
In this section, we report the p-values and other results for our experiments.

\subsection{Source-Dependent Essay Set's p-values}

The results shown here in Table \ref{Seen Essay P-Values} are the p-values for the different essay sets with and without gaze from Table \ref{Results Table}.

\begin{table}[h]
\centering
\begin{tabular}{|l|c|}
\hline
\textbf{Essay Set} & \textbf{p-value} \\
\hline
Prompt 3 & 0.0042 \\
Prompt 4 & 0.0109 \\
Prompt 5 & 0.0133 \\
Prompt 6 & 0.0003 \\
\hline
\end{tabular}
\caption{Source-Dependent essay set's p-values}
\label{Seen Essay P-Values}
\end{table}

\subsection{Unseen Essay Set's p-values}

The results shown here in Table \ref{Seen Essay P-Values} are the p-values for the different essay sets with and without gaze from Table \ref{Unseen Prompts Result Table}.

\begin{table}[h]
\centering
\begin{tabular}{|l|c|}
\hline
\textbf{Essay Set} & \textbf{p-value} \\
\hline
Prompt 1 & 0.0887 \\
Prompt 2 & 0.1380 \\
Prompt 7 & 0.0393 \\
Prompt 8 & 0.0315 \\
\hline
\end{tabular}
\caption{Unseen Essay's p-values}
\label{Uneen Essay P-Values}
\end{table}

\subsection{Native Gaze vs. No Gaze \& All Gaze p-values}

The results shown in Table \ref{No gaze vs. native gaze P-Values} are the p-values for the essay sets using the gaze behaviour of a native English speaker compared to not using gaze behaviour, and using gaze behaviour of all readers.

\begin{table}[h]
\centering
\begin{tabular}{|l|c|c|}
\hline
\textbf{Essay Set} & \textbf{No vs. Native} & \textbf{Native vs. All} \\
\hline
Prompt 1 & 0.1407 & 0.0471 \\
Prompt 2 & 0.0161 & 0.9161 \\
Prompt 3 & 0.3239 & 0.0239 \\
Prompt 4 & 0.0810 & 0.0805 \\
Prompt 5 & 0.4971 & 0.4010 \\
Prompt 6 & 0.2462 & 0.2961 \\
Prompt 7 & 0.0189 & 0.0098 \\
Prompt 8 & 0.8768 & 0.0068 \\
\hline
\end{tabular}
\caption{No gaze vs. native gaze and native gaze vs. all gaze p-values.}
\label{No gaze vs. native gaze P-Values}
\end{table}

\end{document}